\documentclass[letterpaper, 10 pt, conference]{ieeeconf}

\IEEEoverridecommandlockouts                              % This command is only needed if 
\usepackage{hyperref}
\hypersetup{
    colorlinks=false,
    linkcolor=blue,
    filecolor=magenta,      
    urlcolor=blue,
    citecolor=blue,
}
\usepackage{graphics} % for pdf, bitmapped graphics files
\usepackage{epsfig} % for postscript graphics files
\usepackage{mathptmx} % assumes new font selection scheme installed
\usepackage{times} % assumes new font selection scheme installed
\usepackage{amsmath} % assumes amsmath package installed
\usepackage{amssymb} % assumes amsmath package installed
\usepackage{dsfont}
\usepackage{url}
\usepackage{multirow}
\usepackage{graphicx}
\usepackage{cite}

\usepackage{comment}

\title{\LARGE \bf Enhancing the Vertical Mobility of a Robot Hexapod Using Microspines}

\author{Matt Martone$^{*1}$, Catherine Pavlov$^{*2}$, Adam Zeloof$^{2}$, Vivaan Bahl$^{3}$, and Aaron M. Johnson$^{2}$%
\thanks{$^*$M.~M.\ is supported by a DOE Traineeship in Robotics. C.~P.\ is supported by a NASA Space Technology Research Fellowship. Both contributed equally to the work presented here. }%
\thanks{$^{1}$Robotics Institute, Carnegie Mellon University, Pittsburgh, PA, USA {\tt\small mmartone@andrew.cmu.edu}}%
\thanks{$^{2}$Mechanical Engineering, Carnegie Mellon University, Pittsburgh, PA, USA {\tt\small \{cpavlov,azz,amj1\}@andrew.cmu.edu}}%
\thanks{$^{3}$Electrical \& Computer Engineering, Carnegie Mellon University, Pittsburgh, PA, USA {\tt\small vrbahl@andrew.cmu.edu}}%
}

\begin{document}
\maketitle

\begin{abstract}Modern climbing robots have risen to great heights, but mechanisms meant to scale cliffs often locomote slowly and over-cautiously on level ground. Here we introduce T-RHex, an iteration on the classic cockroach-inspired hexapod that has been augmented with microspine feet for climbing. T-RHex is a mechanically intelligent platform capable of efficient locomotion on ground with added climbing abilities. The legs integrate the compliance required for the microspines with the compliance required for locomotion in order to simplify the design and reduce mass. The microspine fabrication is simplified by embedding the spines during an additive manufacturing process. We present results that show that the addition of microspines to the T-RHex platform greatly increases the maximum slope that the robot is able to statically hang on (up to a $45^\circ$ overhang) and ascend (up to $55^\circ$) without sacrificing ground mobility.
\end{abstract}

%\vspace{-10pt}

%%%%%%%%%%%%%%%%%%%%%%%%%%%%%%%%%%%%%%%%%%%%%%%%%%%%%%%%%%%%%%%%%%%%%%%%%%%%%%%%
\section{Introduction}

In nature, animals have adapted a wide variety of approaches to climbing, but even with finely-tuned sensorimotor cortices most can't measure up to the reliability of insects.  Cockroaches are able to scale nearly any natural material using microscopic hairs along their legs and feet for surface adhesion \cite{Dai2479,Beutel2001,Casteren2010}. These hairs catch on surface irregularities, known as asperities, and can support the insect's body through load sharing across many hairs. The cockroach's scant cognitive intelligence severely limits its ability to plan actions or move dexterously; rather it relies on mechanical intelligence for locomotion. This approach is promising for field robots that need to reliably climb on an array of surface materials and geometries.  

\begin{figure}[t]
\centering
%\captionsetup{justification=centering}
\includegraphics[width=\linewidth]{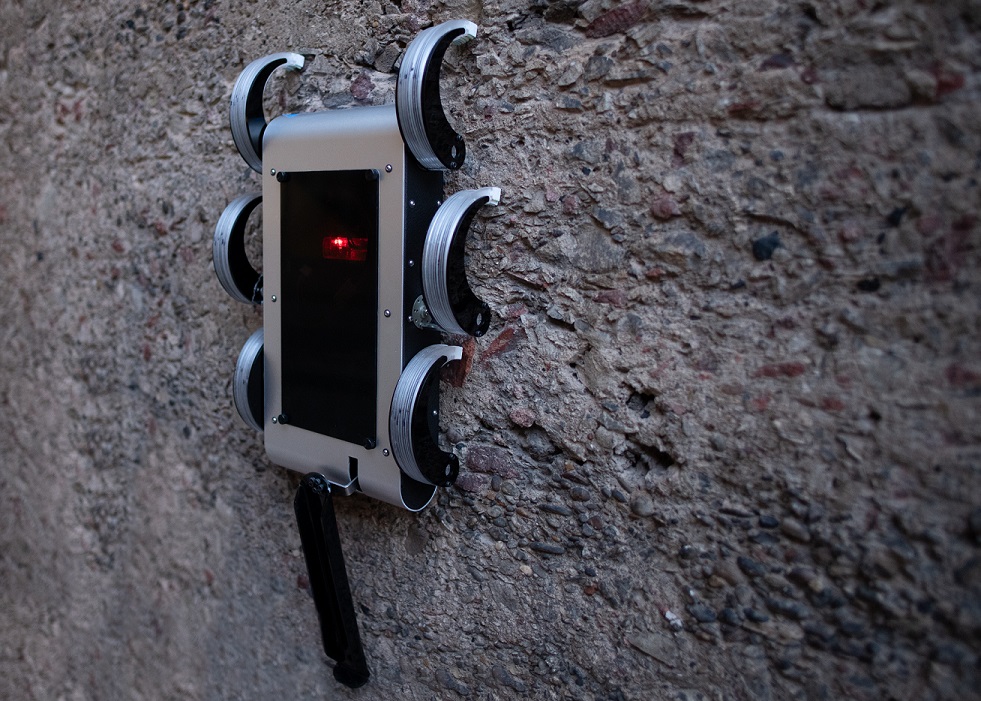}
\caption{The new T-RHex robot platform statically clinging to rough aggregate concrete. Each leg includes 9 individually compliant microspines.}
\label{TRHex}
\vspace{-16pt}
\end{figure}

Microspine technology can effectively mimic the adhesion of a cockroach's foot on a larger scale \cite{Asbeck2005}, though heavily directional without the aid of Van der Waals forces (as used in gecko-like adhesives \cite{Geim2003}). Microspine grippers can attach to rough surfaces by distributing the load across hundreds of millimeter scale hooks sprung with hierarchical compliance. Legged robots such as RiSE \cite{Spenko2008}, Spinybot \cite{Asbeck2006}, and LEMUR \cite{Parness2017} have demonstrated the effectiveness of microspines for robotic rock climbing. However, while these systems can scale walls efficiently, their
microspine feet and slow joints limit mobility on flat ground.
%is limited by slow moving joints and deliberate foot placement. 
%No legged system has been developed to utilize microspines for climbing while retaining dynamic ground agility.
%These systems also take great care to avoid ground contact with any part of  the flexure or cassette assembly other than the spine tips.  The molding process used often results in fragile toes that must be carefully loaded to avoid failure.  

Microspines have been utilized to achieve climbing with numerous morphologies, though all must make design compromises in order to climb. Some like Spinybot \cite{Asbeck2005} are designed to only work in scansorial locomotion. LEMUR3 is a quadruped which uses active microspine grippers on its feet to achieve mobility on vertical and overhung natural rock surfaces, but requires carefully planned foot placement for each step that limits travel speed \cite{Parness2017,Parness2013}. DROP \cite{Carpenter2016} and TBot \cite{liu2015wheeled} are small, lightweight wheeled platforms capable of both vertical and level ground mobility, but microspine adhesion quickly degrades as the tips dull on flat ground. RiSE \cite{Spenko2008} is a hexapedal robot with microspine feet capable of 0.25 m/s locomotion on flat ground, but uses different feet for walking and climbing. 
%As yet, no robot has demonstrated reliable rough terrain locomotion and untethered high angle mobility in a single package.

RHex is a cockroach-inspired ground robot that uses six legs to traverse uneven terrain \cite{OGRHex,Altendorfer2001,xRhex}. Its curved leg shape and compliance are key to simplifying control architecture through mechanical intelligence \cite{koditschek2004mechanical,spagna2007distributed}, and it can move effectively on a wide array of structured and unstructured terrains \cite{Moore2002,li2009sensitive,chou2012bio,paper:ilhan_hill_2018}. RHex can carry a relatively large payload due to the structure of the legs \cite{paper:xrhex_canid_spie_2012}, enabling it to carry many sensors for search and rescue, reconnaissance, or environmental monitoring missions \cite{tr:desert-2014}. While quite mobile on horizontal terrain, the platform's utility and viability would be improved greatly by adding the ability to scale slopes and walls. This is especially important in areas with incomplete infrastructure such as disaster zones or construction sites. 

In this paper we present a new leg design (in Sec.~\ref{sec:legdesign}) that uses microspines to enable a RHex-like robot to hang onto vertical or even overhanging surfaces, such as the vertical wall in Fig.~\ref{TRHex}, as well as climb inclined surfaces up to 55$^\circ$. 
%Curved fiberglass RHex legs are characteristically flexible, deflecting under the robot's weight when in contact with the ground. 
The microspine toes, Sec.~\ref{sec:microspinedesign}, are built using additive manufacturing to simplify the fabrication process.
We leverage the existing leg compliance to avoid the need for multiple smaller flexures to enable successful adhesion (Sec.~\ref{sec:flexure}). By orienting the spines backwards, we ensure that the existing RHex forward gaits are unimpeded while taking advantage of the microspines for climbing with the typically unused backward walking gait space (Sec.~\ref{sec:gait_design}). We test, in Sec.~\ref{sec:experiments}, the new leg design's ability to climb and hang on a variety of inclined surfaces while maintaining good level-ground mobility. Finally, we conclude the paper in Sec.~\ref{sec:future_work} by considering a pathway towards full horizontal and vertical mobility with the platform.

%\begin{comment}
%\section{Motivation}\label{sec:motivation}
%One of the original intents of the RHex platform was for robotic reconnaissance and small payload delivery through rugged environments \cite{OGRHex,Altendorfer2001}. Its simple, robust architecture makes it ideal for locomotion through unstructured environments at relatively high speed for a legged system. While much work has been done to allow RHex to survive falls, leap over gaps, climb stairs, and scale obstacles approximately one body length high, RHex's mobility is largely limited to horizontal terrain \cite{Johnson2013,Moore2002}. The platform's utility and viability when deployed in an urban environment would be improved greatly by adding the ability to scale slopes and walls. This is especially important in areas with incomplete infrastructure such as disaster zones or construction sites. 
%\end{comment}

\section{Mechanical Design}\label{sec:prelimDesign}
The robot described in this paper is an implementation of the standard RHex configuration named ``T-RHex,'' shown in Fig. \ref{TRHex}. Like RHex, T-RHex is a hexapedal robot with six single degree-of-freedom compliant semicircular legs. Each leg is actuated by a Dynamixel MX-64 servo motor capable of continuous $360^\circ$ rotation. Unlike the standard RHex platform, the legs are comprised of multiple independently sprung microspine flexures, whose design is discussed in the rest of this section. The fully assembled T-RHex platform weighs 2.5kg and measures 254mm between front and back legs.

In addition to the new legs, T-RHex is equipped with a ``climbing tail''.
The tail is mounted to what is the front of the robot during regular walking, which becomes the rear of the robot during climbing.
Past tails on RHex have been primarily oriented upwards and towards the rear (while walking) \cite{paper:libby_tail_2016}, and work by using inertial reorientation. The tail on T-RHex is lightweight and intended to contact the surface in order to increase the adhesion forces a the front toes, as with the tail in other climbing robots \cite{autumn2005robotics,kim2005spinybotii,liu2015wheeled}.
The tail is actuated with a Dynamixel MX-106 servo motor and can rotate $180^\circ$.  

\begin{figure}
%\captionsetup{justification=raggedright,singlelinecheck=false}
%\begin{subfigure}
    \centering
    \includegraphics[width=\linewidth]{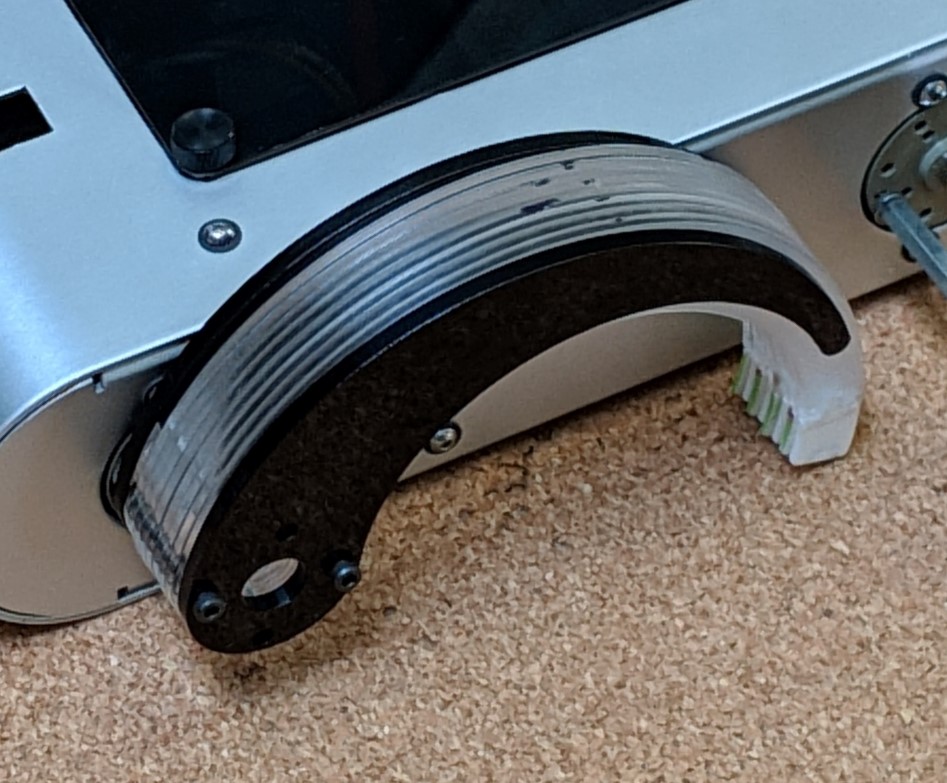}
%        \label{fig:fwhegwall}
%    \vspace{-1.25cm}
%    \caption*{\hspace{.15cm}\textbf{\colorbox{white}{\textcolor{red}{a)}}}}
%    \vspace{.5cm}
%\end{subfigure}%
%\vspace{-6pt}
%\begin{subfigure}
%    \centering
    \includegraphics[width=\linewidth]{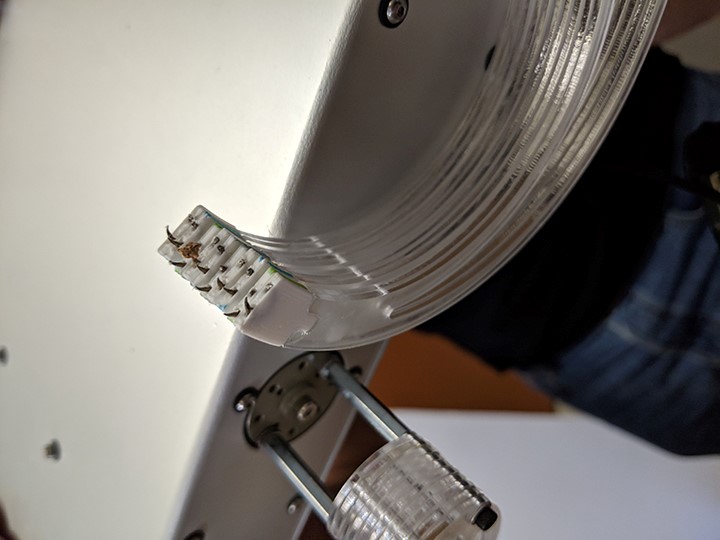}
%        \label{fig:corkfwhegs}
%    \vspace{-1.25cm}
%    \caption*{\hspace{.15cm}\textbf{\colorbox{white}{\textcolor{red}{b)}}}}
%\end{subfigure}
%\captionsetup{justification=centering}
%\vspace{6pt}
\caption{A fully assembled T-RHex forked wheel leg (top) and a closeup of the microspines on the toes (bottom). Both images show a variety of spine angles, denoted by the color on the end of each leg slice.}
\label{fig:fwheg}
\vspace{-11pt}
\end{figure}

\subsection{Leg Design}
\label{sec:legdesign}
RHex's legs were redesigned to enable T-RHex to climb on high-angle terrain. While climbing, T-RHex faces backwards, such that the tips of its appendages are the only points in contact with the surface (an orientation used by some other legged robots with semicircular leg designs \cite{hoover2010bio}). A T-RHex leg consists of multiple stacked, thin slices, as seen in Fig. \ref{fig:fwheg}. Each individual leg slice, Fig.~\ref{fig:leg}, has a single microspine, and the stacked slices are able to deflect in plane independently of each other, with their shared attachment point concentric with the driving servo horn. The thickness of the slices was determined by microspine size, as each slice had to fully enclose its microspine leaving only the tip exposed.

Because the microspines protrude only from the tip of each leg, they do not interfere with the ground during normal forward walking. This means that ground mobility is not impacted, as evaluated in Sec.~\ref{sec:groundmobility}, and the robot can travel at the same speed regardless of whether the legs contain spines or not. This is also useful to prevent surface harm while the robot walks, and to protect the spines from dulling when not in use.

\subsection{Microspine Design}
\label{sec:microspinedesign}

\begin{figure}
%\captionsetup{justification=centering}
\centering
\includegraphics[width=.9\linewidth]{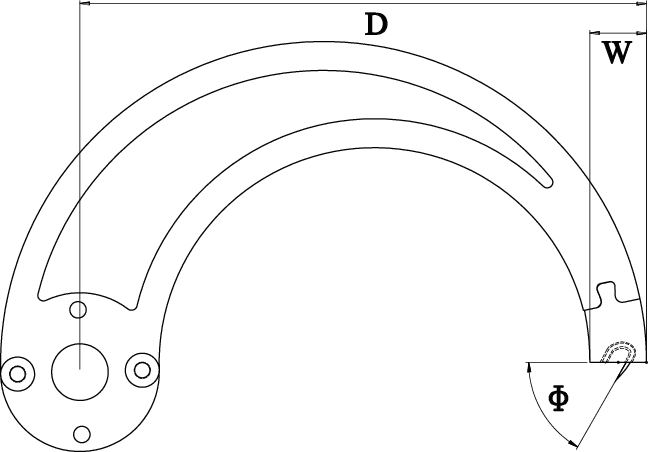}
\caption{Drawing of a single microspine flexure. On T-RHex, D is 100mm, W is 10 mm, and $\phi$ is $60^\circ$, $90^\circ$, or $120^\circ$. A full leg contains nine stacked microspine flexures. }
\label{fig:leg}
\vspace{-11pt}
\end{figure}

Most prior microspine assemblies were fabricated through a multistep additive and subtractive process of polymer casting, milling, and spine embedding called
shape deposition manufacturing (SDM) \cite{merz1994shape,weiss1997shape,binnard2000design}. Here, we take a different approach that uses 3D printing to simplify the fabrication process.
Each microspine is attached to the leg slice with a 3D printed PLA piece. These slice ``toes'' were 3D printed in large batches, and paused mid-print to insert the microspines, which were then printed over. Three different spine angles ($\phi$ in Fig. \ref{fig:leg}) were fabricated: $60^\circ$, $90^\circ$, and $120^\circ$. Each spine angle allows for adhesion in a different range of the leg's rotation, and as such each leg has a mix of spine angles. Unlike prior legged microspine climbers, which keep the angle of attack of the microspines very constant while in stance, T-RHex must maintain adhesion over a large angular range, making this mix of toe angles key to the success of the climbing gait.  

The microspines on T-RHex consist of size 12 plain shank fish hooks cut to length. At approximately 0.6mm thick, these hooks are small enough to allow for 1.5mm thick leg slices. These hooks were chosen for their thickness after preliminary experimentation revealed no discernible effect of hook size on adhesion within the considered group.

\subsection{Flexure Design}
\label{sec:flexure}
Unlike prior microspine systems, which use SDM with multiple materials to tune the compliance of the flexures, we have simplified the design to use a monolithic flexure per spine that meets our compliance needs and is easy to reproduce quickly.  This is similar to the method used by the Asteroid Redirect Mission Gripper 2.0 \cite{ARMGripper}, but improves the repeatability of inserting the spine at the desired angle.  The main portion of each leg slice is constructed of lasercut high-impact acrylic, which is attached to the 3D printed tip with acrylic adhesive on a mechanically interlocking shape (Fig.~\ref{fig:leg}). 

For effective climbing, the microspines must be able to independently translate while maintaining a fixed angle, due to the narrow range of attack angles for adhesion of microspines. The geometry of the leg slice flexure allows the microspine at the tip to deflect on the order of millimeters both parallel and perpendicular to the climbing surface with minimal rotation of the spine. A full leg stack of flexures has a spring constant of 4-5 kN/m, compared to a typical spring constant of 0.9 kN/m on comparably sized RHex legs. 

\subsection{Cassette Design}
The full assembly of each leg is a cassette of nine stacked leg slices and rigid outer plates with thin spacers separating each slice to allow independent movement. The rigid outer walls, shown in black in Fig.~\ref{fig:fwheg} (top), prevent out-of-plane bending by the leg slices but do not come into contact with the surface during walking or climbing. As the legs are spaced slightly farther than one leg length apart, the central legs are spaced outward with standoffs to prevent interference. Each leg is directly bolted to its servo horn. 

\section{Gait Design}\label{sec:gait_design}
A new gait, termed the ``inchworm,'' was developed for wall climbing to accommodate the mode of action of the microspines. This gait is most similar to the metachronal gait used to climb stairs \cite{Moore2002} where pairs of legs move together to propel the robot upwards. Step angles were chosen to maximize the preload force to set the spines as the adhesion force of microspines in shear is roughly proportional to preload force normal to the surface.

A metachronal gait also provides symmetric supporting forces on the left and right sides to avoid producing a net torque about any foot. Asymmetric gaits, such as a backwards alternating tripod, failed quickly because the microspines cannot resist torque in the plane of the wall.

\begin{figure}[t]
    \centering
    \includegraphics[width=\linewidth]{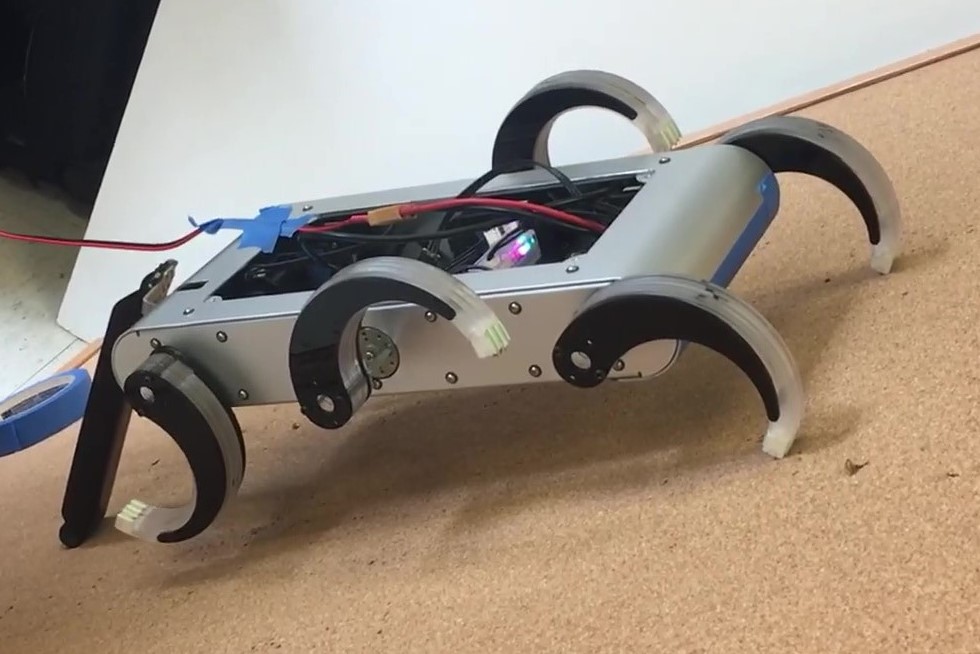}
    \caption{The ``inchworm'' climbing gait in action on a $55^\circ$ slope showing the three pairs of legs moving symmetrically. In this frame the front legs (right) are holding position until the back legs (left) complete a clockwise rotation, and the tail counteracts the tipping torque generated by the microspine preload force.  }
    \label{fig:climbing_gait}
\end{figure}

During the ``inchworm'' gait the back legs provide propulsion, the front legs engage to hold position between steps, and the tail prevents the back legs from catching or leveraging the body off of the wall while recirculating. The central leg pair is not utilized in the gait variation used to generate the results below, but can duplicate the role of the front legs on steeper surfaces. Shown in Fig.~\ref{fig:climbing_gait} is a still frame of T-RHex executing this gait. In the frame, the back legs (left) are beginning to recirculate, while the front legs (right) engage to prevent the robot from sliding back down. The tail allows the back legs to rotate freely without contacting the surface.

This gait requires some tuning for different slope angles and surfaces. For instance, steeper angle climbs rely much more heavily on the balance of the tail than shallower angles do. Vertical ascents, however, suffer from ``vaulting,'' where the robot attempts to adhere the front toes onto the surface, but instead pushes the robot away from the wall, disengaging the back toes and causing a fall.  

\section{Experimental Results}\label{sec:experiments}
%Testing was conducted in several stages to improve climbing performance and validate robot design. Initial trials to determine hook choice, leg shape, spine attachment method, and tine spacing gave clear results later used to justify design choices, the details of which will be here omitted for brevity. The test suites described below were conducted using the completed T-RHex platform with leg shape and body design held constant. 

Testing was largely performed on three surface types, chosen to be representative of the various terrains that a small legged robotic platform might encounter during deployment in an urban environment. The robot's climbing ability was tested extensively on cork board, brick facade, and plywood. 
%Without a specific deployment goal to dictate target materials, and with limited testing time available preventing an exhaustive study, 
These surfaces were selected as a minimum subset of surfaces needed to characterize microspine performance on a generalized exploration scenario. 

\begin{figure}[tb]
%\captionsetup{justification=centering}
\centering
\includegraphics[width=\linewidth]{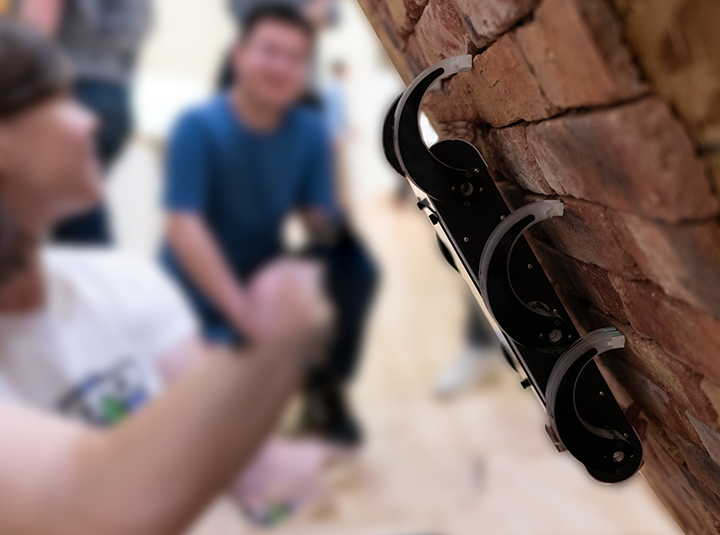}
\caption{T-RHex hanging from brick facade with a steep overhang. The maximum static cling angle recorded during experiments was $116^\circ$ and reached a further $135^\circ$ when placed by hand.}
\label{fig:staticHang}
\end{figure}

The three surface choices represent the three most common failure modes of microspine climbing feet, as described in \cite{shoutouttomyself}. Cork board serves as a soft, pseudo-granular material that microspines can easily puncture and tear through. The expected failure mode is surface degradation due to overloading spines. Plywood is a fairly smooth surface on which microspines must rely equally on existing asperities and active surface deformation by digging into the medium-hardness wood. The failure mode is a lack of strong adhesion points that prevents the microspines from finding purchase. Brick facade serves as a hard, pitted surface that is nearly ideal for microspines to catch strong, favorably-formed asperities. The failure mode of brick facade is likely tine fracture or plastic deformation of the fish hook when feet do not successfully disengage from unusually strong footholds. Fig.~\ref{fig:climbing_gait} and \ref{fig:tipHang} show the cork board and  Fig.~\ref{fig:staticHang} shows the brick test surfaces used in experiments.

\subsection{Static Slope Cling}
\label{sec:static}
The first suite of tests were designed to determine the effectiveness of the T-RHex leg and microspine design without conflating the effects of the climbing gait controller. The leg angle, spine content, and surface material were varied independently while determining the maximum angle at which the robot could statically cling to the wall. The test procedure is as follows:
\begin{enumerate}
    \item Set the position of all the robot's legs to a given angle and command the motors to hold that position indefinitely. 
    \item Place the robot at a randomly selected position on the surface while the surface is held horizontal. The robot must be oriented such that the front while climbing is angled towards the highest point on the surface when raised. 
    \item Rotate the test wall about the bottom edge at a rate of approximately 1 degree per second. This rate is sufficiently quasi-static such that inertial effects of motion can be ignored. 
    \item Carefully observe the robot while clinging to note unusual behaviors, interesting microspine adhesion properties, instances of slip-catch motion, and signs of leg damage. 
    \item Continue pitching the test wall until the microspines fail and the robot falls. Note the angle of the wall when catastrophic detachment occurs. 
\end{enumerate}

The data, listed in Table~\ref{table:cling} and Fig.~\ref{fig:spineData}, was collected across three test surfaces and two leg angles with a sample size of five tests each. As a control, the tests using cassettes of mixed angle microspines were compared to a set of tests using blank legs fabricated as tines without embedded microspines. All statistical comparisons discussed below use a one-tailed, heteroscedastic, t-test. 

\begin{table}[t]
\centering
\caption{Maximum Static Cling Angle Data}
\label{table:cling}
\begin{tabular}{|c|c|c|c|c|c|}
\hline
Tail & Leg Type & Surface & $\theta_{leg}\ (^\circ)$ & $\mu\ (^\circ)$ & $\sigma\ (^\circ)$ \\ \hline
\multirow{12}{*}{No Tail} &\multirow{6}{*}{Microspine} & \multirow{2}{*}{Plywood}    & -150 & 36.2 & 1.10  \\ \cline{4-6} 
                                  & &                            & -120 & 58   & 2.12  \\ \cline{3-6} 
                                  & & \multirow{2}{*}{Cork Board} & -150 & 52   & 3.61  \\ \cline{4-6} 
                                  & &                            & -120 & 79.2 & 3.77  \\ \cline{3-6} 
                                  & & \multirow{2}{*}{Brick}      & -150 & 52.6 & 10.90 \\ \cline{4-6} 
                                  & &                            & -120 & 73.8 & 18.24 \\ \cline{2-6}
& \multirow{6}{*}{Blank}        & \multirow{2}{*}{Plywood}    & -150 & 32.2 & 3.11  \\ \cline{4-6} 
                                  & &                            & -120 & 25.8 & 2.86  \\ \cline{3-6} 
                                  & & \multirow{2}{*}{Cork Board} & -150 & 34.6 & 0.55  \\ \cline{4-6} 
                                  & &                            & -120 & 40   & 2.12  \\ \cline{3-6} 
                                  & & \multirow{2}{*}{Brick}      & -150 & 33.6 & 1.52  \\ \cline{4-6} 
                                  & &                            & -120 & 42   & 5.70  \\ \hline
\multirow{12}{*}{Tail} &\multirow{6}{*}{Microspine} & \multirow{2}{*}{Plywood}    & -150 & 57.2 & 4.60  \\ \cline{4-6} 
                                  & &                            & -120 & 62.4 & 7.50  \\ \cline{3-6} 
                                  & & \multirow{2}{*}{Cork Board} & -150 & 60.4 & 2.70  \\ \cline{4-6} 
                                  & &                            & -120 & 66.4 & 3.44  \\ \cline{3-6} 
                                  & & \multirow{2}{*}{Brick}      & -150 & 69.8 & 21.79 \\ \cline{4-6} 
                                  & &                            & -120 & 92   & 16.31 \\ \cline{2-6}
& \multirow{6}{*}{Blank}        & \multirow{2}{*}{Plywood}    & -150 & 27.4 & 3.21  \\ \cline{4-6} 
                                  &  &                           & -120 & 20.8 & 1.79  \\ \cline{3-6} 
                                  & & \multirow{2}{*}{Cork Board} & -150 & 46.6 & 0.55  \\ \cline{4-6} 
                                  &  &                           & -120 & 40.4 & 0.89  \\ \cline{3-6} 
                                  & & \multirow{2}{*}{Brick}      & -150 & 57.2 & 6.42  \\ \cline{4-6} 
                                  &  &                           & -120 & 43.4 & 12.38 \\ \hline
\end{tabular}
\end{table}

The data indicates that a shallower leg angle of $-120^\circ$ consistently outperformed a steeper leg angle of $-150^\circ$ on the microspine legs ($p < 0.001$), which informed the gait design discussed in Sec.~\ref{sec:gait_design}. Performance was similar between cork board and brick, with plywood underperforming by a margin of about $15^\circ$. However, unlike cork board and plywood, which had tightly clustered data points, the performance on brick, a heterogeneous surface, was highly dependent on initial robot placement, leading to a much higher standard deviation. This is likely due to the variability of asperity quality between brick face and brick edges. Furthermore, the data shows that the spines significantly improved performance of T-RHex over blank legs for the maximum static cling angle ($p < 0.001$). This trend can be seen in Fig.~\ref{fig:spineData}. 

\begin{figure}[t]
%\captionsetup{justification=centering}
\centering
\includegraphics[width=\linewidth]{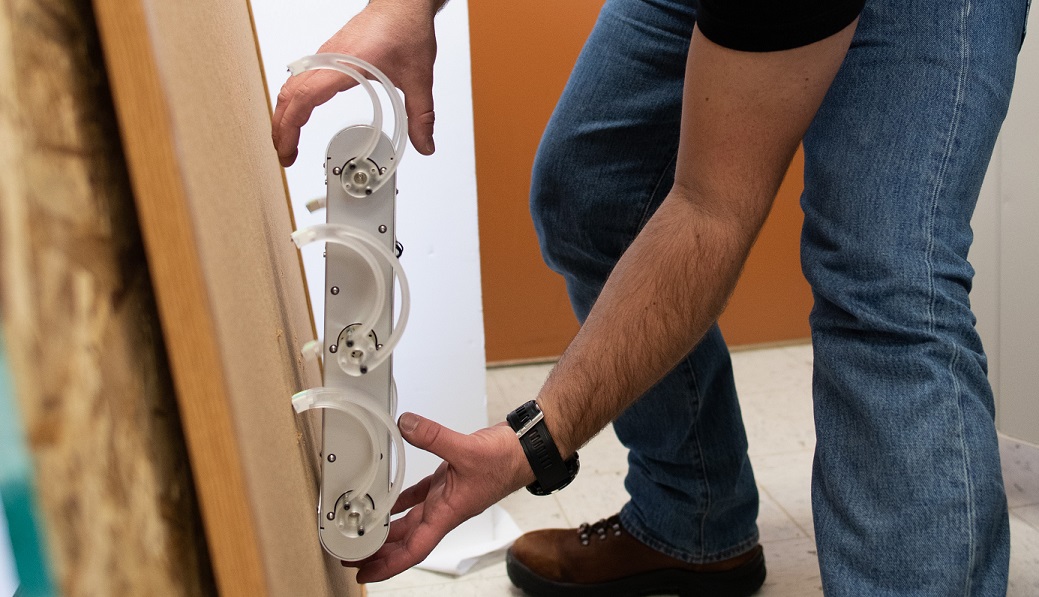}
\caption{A static slope cling test in which the robot is hanging entirely by the back legs after initial tip. The cork test surface is shown here. }
\label{fig:tipHang}
\vspace{-6pt}
\end{figure}

The failure mode of cling was consistently not in the slip of the microspines, but the entire robot body tipping backwards and peeling the spines off of the wall as the center of gravity moved away from the wall as seen in Fig.~\ref{fig:tipHang}. Often, the spines on the front and middle legs would be centimeters away from the wall with the back toes holding all of the robot's weight after reaching an initial tipping angle of around $35^\circ$. This motivated the addition of a tail to T-RHex. 

\subsection{Static Slope Cling with Tail}
We hypothesized that with a smooth tail providing a preload force at a point contact behind and below the robot, the front and middle legs would stay attached to the wall for longer, improving the load sharing between all legs and increasing the maximum static cling angle of the robot. 
An active tail made of heat-formed acrylic was added to the robot and the same suite of tests was repeated. We assume that the added mass of the acrylic is negligible compared to T-RHex's total mass. The tail was positioned to be in contact with the horizontal surface at the start of the test. 

\begin{figure}[!ht]
%\captionsetup{justification=centering}
\centering
\includegraphics[width=\linewidth]{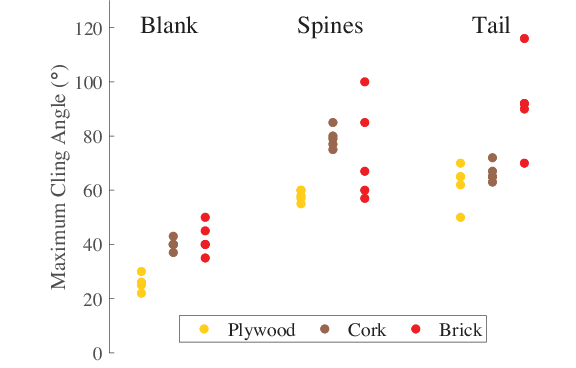}
\vspace{-16pt}
\caption{Maximum static cling angles for the robot with blank (spineless) legs, legs with spines, and with both spines and a tail. All data shown is for a leg angle of $-120^\circ$, and tail data is with legs with tines. Microspines significantly improve the ability to cling at steep angles, but the addition of a tail has a relatively minor impact on static cling angle.}
\label{fig:spineData}
\vspace{-11pt}
\end{figure}

The tailed robot was tested with the same method as in Sec.~\ref{sec:static}, and the results are listed in Table~\ref{table:cling} and Fig.~\ref{fig:spineData}. The tail succeeded in delaying the robot tip angle, but had mixed results on climbing performance. Because the asperity quality of the plywood was poor, the failure occurred as a slip in what could be modeled as high coefficient Coulombic friction, largely independent of number of spines engaged. The addition of a tail did not improve the performance of T-RHex on plywood with reliable statistical significance ($p = 0.1334$), as seen in Fig. \ref{fig:spineData}. On brick, the tail kept more spines engaged to higher angles, which may have improved performance, though with poor statistical significance ($p = 0.0832)$. On cork board, the tail actually hurt performance ($p < 0.001$). We believe that this is due to the mechanism of engagement with a soft surface. Our hypothesis is that with more spines in contact there is less preload force per spine, thus they no longer dig in as far, leading to a poorer grip and earlier fall. 

In all, testing with the tail showed that it is not as beneficial for clinging as we had hoped, but it is extremely reliable as a mechanism to prevent tipping, which is critical for designing gaits that involve considerable pitching of the robot body, Sec.~\ref{sec:gait_design}.

\subsection{Maximum Climb Angle}
Testing was conducted to determine the maximum angle T-RHex could climb. In these experiments, we demonstrated improvement in climbing ability with the addition of microspines. There was slight variation in the microspine leg design between static and climbing tests, with the spines redistributed among legs and rigid sidewalls added after completion of static testing. These modifications did not impact static cling ability, as they served to counter two problems observed during climbing: failure of the microspines to detach and out of plane bending of leg slices during multiple steps. The robot used its tail for all climbing experiments.

Experiments were conducted on the same test surfaces as the static cling tests, with both microspine and blank legs. Fig.~\ref{fig:climbing_gait} shows the robot outfitted with microspine legs during a test. For each test, the robot was started using the climbing gait described in Sec.~\ref{sec:gait_design} and placed on the climbing surface. The surface was held at a fixed angle, and it was determined whether the robot was capable of making observable (1 cm) upward progress over the course of several steps. Binary search was used to determine the maximum angle the robot could climb for each climbing surface and leg combination, with results compiled in Table \ref{table:climbingtests}. Note that microspines greatly outperformed blank legs on all three test surfaces.

\begin{table}[!ht]
\centering
\caption{Maximum Climbing Angle With Tail}
\begin{tabular}{|c|c|c|}
\hline
Leg Type & Surface Material & $\theta\ (^\circ)$ \\ 
\hline
\multirow{3}{*}{Microspine} & {Plywood}    & 30  \\ \cline{2-3}
                                  & {Cork Board} & 50\\ \cline{2-3}
                                  & {Brick}      & 35\\  \hline
\multirow{3}{*}{Blank}        & {Plywood}    & 5\\ \cline{2-3} 
                                  & {Cork Board} & 25 \\ \cline{2-3} 
                                  & {Brick}      & 22 \\  \hline
\end{tabular}
\label{table:climbingtests}
\vspace{-11pt}
\end{table}

\subsection{Flat Surface Mobility}
\label{sec:groundmobility}
A series of tests were conducted to demonstrate the ground mobility of the redesigned legs, and to compare them to legs more typical of RHex platforms. The RHex legs utilized in these tests come from Edubot \cite{xRhex}, a RHex platform of the same scale presented here. 

Mobility tests were conducted on two surfaces: flat pavement and large, smooth rocks approximately the same radius as the legs. The robot was run in an alternating tripod gait three times across a 3 meter section of each surface with each set of legs. Both sets of legs were able to reliably walk on both of these surfaces, demonstrating that the microspine legs are capable of good ground mobility even on highly unstructured terrain. 

On level pavement, the robot's average speed was 0.24 m/s with T-RHex legs and 0.27 m/s with Edubot legs. On the rocks, the robot's speed was 0.16 m/s with T-RHex legs and 0.20 m/s with Edubot legs. The robot was slightly slower on both surfaces with T-RHex legs than Edubot legs, though it walked noticeably straighter with T-RHex legs. The difference in speed is likely due to the presence of rubber treads on the Edubot legs, which reduce slippage during steps. Note that max speed of the robot was limited by the maximum speed of the Dynamixel motors used.

\subsection{Additional Testing}
\begin{figure}[t]
%\captionsetup{justification=centering}
\centering
\includegraphics[width=\linewidth]{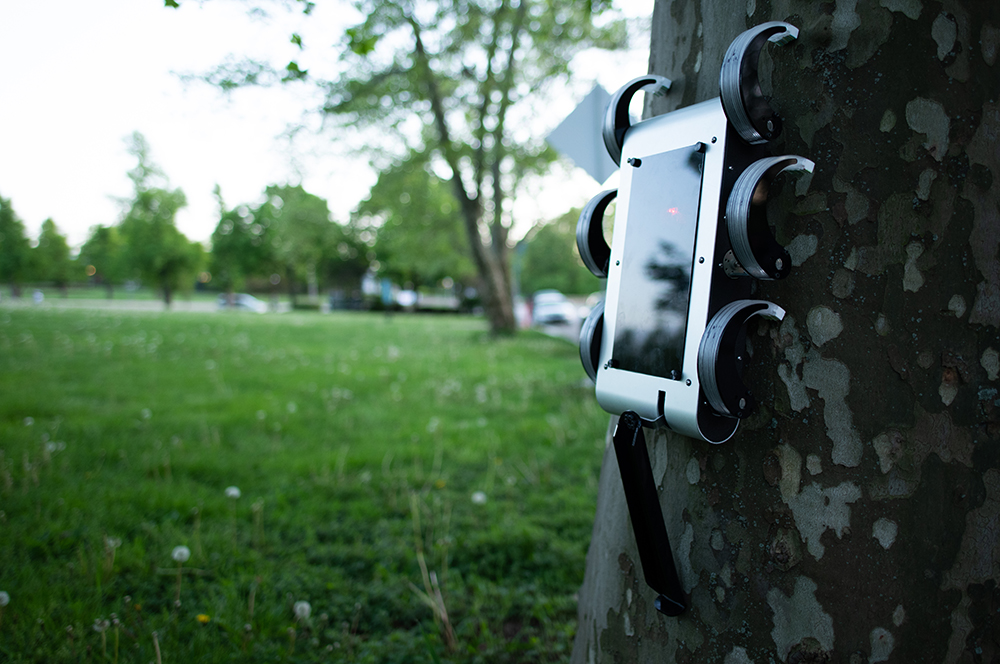}
\caption{The T-RHex platform statically hanging from a Sycamore tree.}
\label{fig:TRHexTree}
\end{figure}

Since the ultimate goal was to climb surfaces outside of the lab with T-RHex, we explored many real-world surfaces on which to test the robot. 
%Our brick facade test surface proved too inconsistent for reliable climbing due to a dearth of asperities on some brick faces. 
The robot was able to statically cling to vertical surfaces of the following materials: Sycamore tree bark (Fig.~\ref{fig:TRHexTree}), cinderblock, fabric bulletin board, coarse aggregate concrete (Fig.~\ref{TRHex}), and padded wall. The highest climbable slope discovered was $55^\circ$ on the textured concrete of a building roof.

Though T-RHex was unable to ascend a vertical wall, it surpassed all expectations with ``best-case'' cling testing. In these trials, performed with the same method as other static cling testing, the robot's starting position was expertly selected on the brick surface as a point that had particularly intriguing asperities. This testing revealed that T-RHex could reliably hang onto the wall with significant overhang past vertical. In some cases, the maximum cling angle reached $135^\circ$, an overhang of $45^\circ$ (steeper than the trial shown in Fig.~\ref{fig:tipHang}). 

\section{Future Work}\label{sec:future_work}

%Our results indicate that it is possible to augment existing robot architectures with microspines to enable climbing and expand the set of traversable terrain. T-RHex is able to statically hang on surfaces overhanging up to $45^\circ$ and climb on surfaces up to $55^\circ$, with little impediment to flat ground walking. The microspine fabrication is greatly simplified by using rapid prototyping techniques and taking advantage of the existing leg compliance. Much like the insects that inspired it, this platform has the potential to become a robot that can truly go anywhere. 

T-RHex is a highly promising prototype, but reaching its full potential requires additional systems work. Future work on refining leg design, enhanced sensing and autonomy, and different modes of control are needed to fully realize autonomous climbing of vertical surfaces.

%\subsection{Exploring Different Leg Materials}
The T-RHex legs are primarily acrylic, a decision made for easy, inexpensive prototyping and rapid design refinement. We did not attempt to match the overall spring constant of RHex's legs, a feature needed for dynamic running. Finding a material that can provide a lower spring constant with equal or greater yield strength will allow for more enhanced mobility. Additional materials considered include fiberglass, spring steel, and more exotic materials such as metallic glass (which has been used in previous microspine flexures \cite{shoutouttomyself}). 

%\subsection{Increased Autonomy}
The onboard software used here only allows for a single gait to be run. For real-world deployment on multi-angled terrains of varying surface material, T-RHex requires increased autonomy and the ability for operators to modify gaits without losing power. 
%This comes in two major forms: the floor-to-wall transition and in-situ gait switching.

In particular, the floor-to-wall transition is essential for full autonomy.  The robot should be able to recognize a climbable surface, and automatically position itself at the base of the slope or wall such that the microspines can engage and employ the wall-climbing gait.  Achieving this functionality requires scripting a behavior for the transition and adding the ability to use onboard vision or inertial sensing to recognize when to perform the behavior. The transition may also require hardware changes such as adding a body pitch degree of freedom \cite{Spenko2008}.

%\subsection{Sensing and Closed Loop Control}
As of now, T-RHex moves with an entirely open-loop, scripted gait. We believe that performance can be improved by adding torque sensing to the leg and tail actuators, which would allow the robot to adapt to large-scale surface irregularities in the wall, detect leg slip before catastrophic detachment, and automatically use the tail to balance during wall climbs. This design path would require a platform overhaul, but offers a promising controls-based solution to the shortcomings of an open-loop gait. 

\section{Conclusion}\label{sec:conclusion}
Our results indicate that it is possible to augment existing robot architectures with microspines to enable climbing and expand the set of traversable terrain.  Our robot, T-RHex, was able to statically hang on surfaces overhanging up to $45^\circ$, and climb on surfaces up to $55^\circ$, with little impediment to flat ground walking. The T-RHex platform will benefit from exploration of additional leg materials, as well as further gait tuning. Much like the insects that inspired it, this platform has the potential to become a robot that can truly go anywhere. 

\bibliographystyle{IEEEtran}
\bibliography{reference.bib}

\begin{thebibliography}{10}
\providecommand{\url}[1]{#1}
\csname url@samestyle\endcsname
\providecommand{\newblock}{\relax}
\providecommand{\bibinfo}[2]{#2}
\providecommand{\BIBentrySTDinterwordspacing}{\spaceskip=0pt\relax}
\providecommand{\BIBentryALTinterwordstretchfactor}{4}
\providecommand{\BIBentryALTinterwordspacing}{\spaceskip=\fontdimen2\font plus
\BIBentryALTinterwordstretchfactor\fontdimen3\font minus
  \fontdimen4\font\relax}
\providecommand{\BIBforeignlanguage}[2]{{%
\expandafter\ifx\csname l@#1\endcsname\relax
\typeout{** WARNING: IEEEtran.bst: No hyphenation pattern has been}%
\typeout{** loaded for the language `#1'. Using the pattern for}%
\typeout{** the default language instead.}%
\else
\language=\csname l@#1\endcsname
\fi
#2}}
\providecommand{\BIBdecl}{\relax}
\BIBdecl

\bibitem{Dai2479}
Z.~Dai, S.~N. Gorb, and U.~Schwarz, ``Roughness-dependent friction force of the
  tarsal claw system in the beetle {Pachnoda} marginata ({Coleoptera},
  {Scarabaeidae}),'' \emph{Journal of Experimental Biology}, vol. 205, no.~16,
  pp. 2479--2488, 2002.

\bibitem{Beutel2001}
R.~G. Beutel and S.~N. Gorb, ``{Ultrastructure of attachment specializations of
  hexapods (Arthropoda): evolutionary patterns inferred from a revised ordinal
  phylogeny},'' \emph{Journal of Zoological Systematics and Evolutionary
  Research}, vol.~39, pp. 177--207, 2001.

\bibitem{Casteren2010}
A.~van Casteren and J.~R. Codd, ``{Foot Morphology and Substrate Adhesion in
  the Madagascan Hissing Cockroach, Gromphadorhina portentosa},'' \emph{Journal
  of Insect Science}, vol.~10, 2010.

\bibitem{Asbeck2005}
A.~T. Asbeck, S.~Kim, W.~R. Provancher, M.~R. Cutkosky, and M.~Lanzetta,
  ``Scaling hard vertical surfaces with compliant microspine arrays,'' in
  \emph{Robotics:Science and Systems}, June 2005.

\bibitem{Geim2003}
A.~K. Geim, S.~V. Dubonos, I.~V. Grigorieva, K.~S. Novoselov, A.~A. Zhukov, and
  S.~Y. Shapoval, ``{Microfabricated adhesive mimicking gecko foot-hair},''
  \emph{Nature Materials}, vol.~2, no.~7, pp. 461--463, 2003.

\bibitem{Spenko2008}
M.~J. Spenko, G.~C. Haynes, J.~A. Saunders, M.~R. Cutkosky, A.~A. Rizzi, R.~J.
  Full, and D.~E. Koditschek, ``{Biologically Inspired Climbing with a
  Hexapedal Robot},'' \emph{Journal of Field Robotics}, vol.~25, no. 4-5, pp.
  223--242, 2008.

\bibitem{Asbeck2006}
A.~Asbeck, S.~Kim, A.~McClung, and A.~Parness, ``{Climbing walls with
  microspines},'' in \emph{IEEE International Conference on Robotics and
  Automation}, May 2006.

\bibitem{Parness2017}
A.~Parness, N.~Abcouwer, C.~Fuller, N.~Wiltsie, J.~Nash, and B.~Kennedy,
  ``{LEMUR} 3: {A} limbed climbing robot for extreme terrain mobility in
  space,'' in \emph{IEEE International Conference on Robotics and Automation},
  2017, pp. 5467--5473.

\bibitem{Parness2013}
A.~Parness, M.~Frost, N.~Thatte, J.~P. King, K.~Witkoe, M.~Nevarez, M.~Garrett,
  H.~Aghazarian, and B.~Kennedy, ``Gravity-independent rock-climbing robot and
  a sample acquisition tool with microspine grippers,'' \emph{Journal of Field
  Robotics}, vol.~30, no.~6, pp. 897--915, 2013.

\bibitem{Carpenter2016}
K.~Carpenter, N.~Wiltsie, and A.~Parness, ``{Rotary Microspine Rough Surface
  Mobility},'' \emph{IEEE/ASME Transactions on Mechatronics}, vol.~21, no.~5,
  pp. 2378--2390, 2016.

\bibitem{liu2015wheeled}
Y.~Liu, S.~Sun, X.~Wu, and T.~Mei, ``A wheeled wall-climbing robot with
  bio-inspired spine mechanisms,'' \emph{Journal of Bionic Engineering},
  vol.~12, no.~1, pp. 17--28, 2015.

\bibitem{OGRHex}
U.~Saranli, M.~Buehler, and D.~E. Koditschek, ``{RHex}: A simple and highly
  mobile hexapod robot,'' \emph{The International Journal of Robotics
  Research}, vol.~20, no.~07, pp. 616--631, 2001.

\bibitem{Altendorfer2001}
R.~Altendorfer, N.~Moore, H.~Komsuoglu, M.~Buehler, H.~{Brown Jr.},
  D.~McMordie, U.~Saranlie, R.~Full, and D.~Koditschek, ``{RHex: A
  Biologitcally Inspired Hexapod Runner},'' \emph{Autonomous Robots}, vol.~11,
  pp. 207--213, 2001.

\bibitem{xRhex}
K.~C. Galloway, G.~C. Haynes, B.~D. Ilhan, A.~M. Johnson, R.~Knopf, G.~Lynch,
  B.~Plotnick, M.~White, and D.~E. Koditschek, ``X-{RHex}: {A} highly mobile
  hexapedal robot for sensorimotor tasks,'' University of Pennsylvania,
  Philadelphia, PA, Tech. Rep., 2010.

\bibitem{koditschek2004mechanical}
D.~E. Koditschek, R.~J. Full, and M.~Buehler, ``Mechanical aspects of legged
  locomotion control,'' \emph{Arthropod structure \& development}, vol.~33,
  no.~3, pp. 251--272, 2004.

\bibitem{spagna2007distributed}
J.~C. Spagna, D.~I. Goldman, P.-C. Lin, D.~E. Koditschek, and R.~J. Full,
  ``Distributed mechanical feedback in arthropods and robots simplifies control
  of rapid running on challenging terrain,'' \emph{Bioinspiration \&
  biomimetics}, vol.~2, no.~1, p.~9, 2007.

\bibitem{Moore2002}
E.~Moore, D.~Campbell, F.~Grimminger, and M.~Buehler, ``Reliable stair climbing
  in the simple hexapod '{RHex}','' in \emph{IEEE International Conference on
  Robotics and Automation}, 02 2002, pp. 2222--2227.

\bibitem{li2009sensitive}
C.~Li, P.~B. Umbanhowar, H.~Komsuoglu, D.~E. Koditschek, and D.~I. Goldman,
  ``Sensitive dependence of the motion of a legged robot on granular media,''
  \emph{Proceedings of the National Academy of Sciences}, vol. 106, no.~9, pp.
  3029--3034, 2009.

\bibitem{chou2012bio}
Y.-C. Chou, W.-S. Yu, K.-J. Huang, and P.-C. Lin, ``Bio-inspired step-climbing
  in a hexapod robot,'' \emph{Bioinspiration \& biomimetics}, vol.~7, no.~3, p.
  036008, 2012.

\bibitem{paper:ilhan_hill_2018}
B.~D. Ilhan, A.~M. Johnson, and D.~E. Koditschek, ``Autonomous legged hill
  ascent,'' \emph{Journal of Field Robotics}, vol.~35, no.~5, pp. 802--832,
  August 2018.

\bibitem{paper:xrhex_canid_spie_2012}
G.~C. Haynes, J.~Pusey, R.~Knopf, A.~M. Johnson, and D.~E. Koditschek,
  ``Laboratory on legs: {An} architecture for adjustable morphology with legged
  robots,'' in \emph{Unmanned Systems Technology XIV}, vol. 8387, no.~1.\hskip
  1em plus 0.5em minus 0.4em\relax SPIE, 2012, p. 83870W.

\bibitem{tr:desert-2014}
S.~Roberts, J.~Duperret, A.~M. Johnson, S.~v. Pelt, T.~Zobeck, N.~Lancaster,
  and D.~E. Koditschek, ``Desert {RHex} technical report: {Jornada} and {White}
  {Sands} trip,'' University of Pennsylvania, Philadelphia, PA, Tech. Rep.,
  2014.

\bibitem{paper:libby_tail_2016}
T.~Libby, A.~M. Johnson, E.~Chang-Siu, R.~J. Full, and D.~E. Koditschek,
  ``Comparative design, scaling, and control of appendages for inertial
  reorientation,'' \emph{IEEE Transactions on Robotics}, vol.~32, no.~6, pp.
  1380--1398, 2016.

\bibitem{autumn2005robotics}
K.~Autumn, M.~Buehler, M.~Cutkosky, R.~Fearing, R.~J. Full, D.~Goldman,
  R.~Groff, W.~Provancher, A.~A. Rizzi, U.~Saranli \emph{et~al.}, ``Robotics in
  scansorial environments,'' in \emph{Unmanned ground vehicle technology VII},
  vol. 5804.\hskip 1em plus 0.5em minus 0.4em\relax International Society for
  Optics and Photonics, 2005, pp. 291--302.

\bibitem{kim2005spinybotii}
S.~Kim, A.~T. Asbeck, M.~R. Cutkosky, and W.~R. Provancher, ``{SpinybotII}:
  {C}limbing hard walls with compliant microspines,'' in \emph{International
  Conference on Advanced Robotics}.\hskip 1em plus 0.5em minus 0.4em\relax
  IEEE, 2005, pp. 601--606.

\bibitem{hoover2010bio}
A.~M. Hoover, S.~Burden, X.-Y. Fu, S.~S. Sastry, and R.~S. Fearing,
  ``Bio-inspired design and dynamic maneuverability of a minimally actuated
  six-legged robot,'' in \emph{2010 3rd IEEE RAS \& EMBS International
  Conference on Biomedical Robotics and Biomechatronics}.\hskip 1em plus 0.5em
  minus 0.4em\relax IEEE, 2010, pp. 869--876.

\bibitem{merz1994shape}
R.~Merz, F.~Prinz, K.~Ramaswami, M.~Terk, and L.~Weiss, ``Shape deposition
  manufacturing,'' in \emph{International Solid Freeform Fabrication
  Symposium}, 1994.

\bibitem{weiss1997shape}
L.~E. Weiss, R.~Merz, F.~B. Prinz, G.~Neplotnik, P.~Padmanabhan, L.~Schultz,
  and K.~Ramaswami, ``Shape deposition manufacturing of heterogeneous
  structures,'' \emph{Journal of Manufacturing Systems}, vol.~16, no.~4, pp.
  239--248, 1997.

\bibitem{binnard2000design}
M.~Binnard and M.~R. Cutkosky, ``Design by composition for layered
  manufacturing,'' \emph{Journal of Mechanical Design}, vol. 122, no.~1, pp.
  91--101, 2000.

\bibitem{ARMGripper}
E.~G. Merriam, A.~B. Berg, A.~Willig, A.~Parness, and T.~Frey, ``Microspine
  gripping mechanism for asteroid capture,'' in \emph{43rd Aerospace Mechanisms
  Symposium}.\hskip 1em plus 0.5em minus 0.4em\relax NASA Ames Research Center,
  May 2016, pp. 401--414.

\bibitem{shoutouttomyself}
M.~Martone, A.~Parness, and A.~Willig, ``Design and testing of the microspine
  gripper tool for the {Asteroid Redirect Mission},'' Caltech SURF Report,
  California Institute of Technology, Tech. Rep., 2016.

\end{thebibliography}
\end{document}